\pdfoutput=1

\documentclass[11pt]{article}

\usepackage[preprint]{acl}

\usepackage{times}
\usepackage{latexsym}
\usepackage{amsmath}
\usepackage{comment}
\usepackage[T1]{fontenc}

\usepackage[utf8]{inputenc}
\DeclareUnicodeCharacter{2009}{\thinspace} 
\DeclareUnicodeCharacter{00A0}{~}          
\DeclareUnicodeCharacter{2212}{-}

\usepackage{microtype}
\usepackage{inconsolata}

\usepackage{graphicx}

\usepackage{booktabs}
\usepackage{tabularx}
\usepackage{adjustbox}   
\newcolumntype{L}{>{\raggedright\arraybackslash}X}

\usepackage{times}
\usepackage{latexsym}
\usepackage{microtype}
\usepackage{adjustbox}
\usepackage{booktabs}
\usepackage{footnote}
\usepackage{subcaption}
\usepackage{url}

\usepackage{booktabs}
\usepackage{graphicx}

\usepackage{makecell}
\usepackage{array}
\usepackage{booktabs} 

\usepackage{inconsolata}
\usepackage{arabtex}
\usepackage{utf8}
\setcode{utf8}

\renewenvironment{abstract}
  { 
    \begin{center}
      \bfseries Abstract
    \end{center}
    \small 
    \quotation 
    \noindent\ignorespaces 
  }
  { 
    \endquotation 
  }

\title{ATHAR: A High-Quality and Diverse Dataset for Classical Arabic to English Translation}

\author{Mohammed Khalil \\
  Independent Researcher \\
  \texttt{mohammed.khalil.mah@gmail.com} \\\And
  Mohammed Sabry \\
  ADAPT/DCU, Dublin, Ireland \\
  \texttt{mohammed.sabry@adaptcentre.ie} \\}

\begin{document}
\maketitle
\begin{abstract}
Classical Arabic represents a significant era that encompasses the golden age of Arab culture, philosophy, and scientific literature. With a broad consensus on the importance of translating these literatures to enrich knowledge dissemination across communities, the advent of large language models (LLMs) and translation systems offers promising tools to facilitate this goal. However, we have identified a scarcity of translation datasets in Classical Arabic, which are often limited in scope and topics, hindering the development of high-quality translation systems. In response, we present the ATHAR dataset, which comprises 66,000 high-quality classical Arabic to English translation samples that cover a wide array of topics including science, culture, and philosophy. Furthermore, we assess the performance of current state-of-the-art LLMs under various settings, concluding that there is a need for such datasets in current systems. Our findings highlight how models can benefit from fine-tuning or incorporating this dataset into their pretraining pipelines. The dataset is publicly available on the HuggingFace Data Hub: \url{https://huggingface.co/datasets/mohamed-khalil/ATHAR}.
\end{abstract}

\section{Introduction}


Classical Arabic is the foundation of Arabic linguistic theory and is well comprehended by educated Arabic readers. It significantly differs from Modern Standard Arabic (MSA) it is also called (Arangiyya~\footnote{In linguistic discourse, the term “Arangiyya” denotes any simplified or colloquial variety of Arabic.}), which is more simplified in terms of its vocabulary, syntax, morphology, phraseology, and semantics.

Classical Arabic poses unique challenges for accurate translation into English. Unlike MSA, which dominates formal speeches, news channels, and modern literary works, and urban dialects prevalent on social media platforms, Classical Arabic is less commonly used today. Yet, it remains vital, present in many historical documents, books, and literary texts rich with knowledge from the Arab and Muslim golden ages, all awaiting translation and broader exposure.

Current translation systems, including Google Translate and large language models like ChatGPT and Llama, struggle with Classical Arabic, often neglecting it in favour of MSA and urban dialects during dataset creation for machine translation.

This work introduces the ATHAR dataset, a translation resource from Classical Arabic to English. “ATHAR” “\<أثر>” means "legacy" or "ancient work." It represents the literary and cultural heritage and underscores the dataset's role in illuminating classical Arabic texts, emphasizing their importance in preserving and conveying this heritage. The ATHAR dataset aims to address the representativeness and quality limitations of previous datasets. 

This work is organised as follows: Section~\ref{sec:related_work} explores the challenges faced by previous researchers in translating Classical Arabic and details how the ATHAR dataset addresses these challenges. Section~\ref{sec:ATHAR_dataset} elaborates on the methodologies used to create the ATHAR dataset, including steps for data collection, cleaning, and preprocessing to ensure the quality and reliability of the data. In Section~\ref{sec:experiments} we conduct experiments to assess the performance of state-of-the-art LLMs on the ATHAR dataset across various settings such as zero-shot, few-shot, and fine-tuning scenarios. The paper concludes with Section~\ref{sec:conclusion}, highlighting the importance of the ATHAR dataset in developing culturally and linguistically authentic Arabic language models and advancing Arabic natural language processing.


\section{Related Work} \label{sec:related_work}
The notable gap in datasets for Classical Arabic has led to several efforts to gather more resources for Arabic Natural Language Processing (NLP). Prominent among these are the Tanzil and Authentic Hadith datasets, which draw from religious texts. The Tanzil dataset offers translations of the Quran in over 40 languages, including Arabic to English, and is hosted on \url{Tanzil.net} and the OPUS database \cite{tiedemann-2012-parallel}. The Authentic Hadith dataset provides translations of the sayings and practices of the Prophet Muhammad, known for its authenticity and rigorous translation process \cite{IJASAT2199}. While these datasets are rich, they mainly focus on religious content and don't fully represent the diverse genres of classical Arabic literature. Additionally, the Poem Comprehensive Dataset (PCD)~\cite{Yousef2019LearningMetersArabicEnglish-arxiv} provides a dataset focused on Classical Arabic poetry. While this dataset is a valuable resource, it encompasses a limited range of thematic areas.

In contrast, there are numerous datasets for Modern Arabic that include a rich and diverse context, such as the OPUS-100 dataset \cite{zhang-etal-2020-improving}, the MultiUN dataset \cite{eisele-chen-2010-multiun}, and the IWSLT2017 dataset \cite{cettolo-etal-2017-overview}. However, Modern Arabic differs significantly from Classical Arabic in its vocabulary, syntax, and stylistic features, which are not well-represented in these contemporary datasets.

Additionally, significant efforts like those by \citet{wrro81836} have focused on Arabic historical linguistics, producing datasets that explore the evolution and contexts of the Arabic language. Although these datasets are not directly applicable in practical translation tasks due to their lack of translations into other languages, they offer invaluable resources for pretraining LLMs with the knowledge necessary to distinguish between Classical and Modern Arabic. Moreover, the initiative by \citet{aloui2024101billionarabicwords} introduced a corpus of 101 billion Arabic words, crucial for developing LLMs targeted at the Semitic Arabic language. This extensive corpus, predominantly in Modern Arabic with some Classical content, could help LLMs understand Classical Arabic, particularly when combined with smaller, specialized downstream translation datasets.

ATHAR dataset aims to address the representativeness issues in previous classical Arabic datasets by compiling sentences from various contexts and historical periods on topics like science, medicine, philosophy, and culture. This dataset will help fill the gaps in classical Arabic resources and provide a more comprehensive foundation for developing effective translation models.

\section{ATHAR Dataset}\label{sec:ATHAR_dataset}

This section outlines the development of the ATHAR dataset. We start by identifying the sources from which the data was collected. Subsequently, we detail the processing steps implemented to ensure the dataset's high quality. Additionally, we compare ATHAR to previous classical Arabic datasets and well-known modern Arabic datasets. In Appendix~\ref{appendix_data_sample}, we showcase samples of the ATHAR datasets.

\subsection{Data Collection}
The \textbf{ATHAR} corpus comprises \textbf{66k} Arabic–English sentence pairs
extracted from 18 seminal works of Classical Arabic, so it is divided into \textbf{65k} for training and \textbf{1k} for testing%
\footnote{At the time of data collection and publication of this work,
there were no restrictions on scraping resources from
\url{https://rasaif.com/}, the public digital library from which we obtained
the raw texts.}
These sources span the \mbox{8\textsuperscript{th}--14\textsuperscript{th}} centuries
and cover a remarkable range of genres: history, travel writing, philosophy,
science, medicine, poetry, \emph{adab}, and more, thus offering broad insight
into medieval Islamic and world intellectual life.
A concise inventory of the 18 works, together with their centuries,
topical domains, and sentence counts, appears in
Table~\ref{tab:athar-sources} (Appendix~\ref{appendix_athar_data_sources}).

\subsection{Preprocessing}\label{subsec:preprocessing}
To prepare the dataset for use in machine translation models, several preprocessing steps were undertaken:


\paragraph{Cleaning the Data:} During the initial stages of the ATHAR dataset collection process, the primary challenge we encountered involved entries where Arabic and English texts were flipped within HTML class labels we estimate their number at around 15\%-20\%. For further details on this issue, see Appendix~\ref{sec:flipped_cell_appendix}. To address this, we implemented a simple rule-based technique that identifies the language of the text based on the predominance of characters from the respective language's alphabet. After collecting the data, we found the texts contained various types of noise such as empty entries, incorrect sentences, duplicate entries, entries consisting solely of numbers, and other unwanted characters. These issues were systematically identified and removed to enhance the dataset's quality. Additionally, unnecessary columns like "book" and "author" were deleted to focus exclusively on the translation pairs. We also removed religious Quranic verses from the dataset, as they were few in number and not dealt with correctly.

\paragraph{Alignment Verification:} As in the Rasaif websites—where we collected the translations from—the translations are created by human volunteers. Given the lack of detailed insights into their methods, and to ensure that each Arabic sentence was correctly aligned with its English translation, thereby maintaining the context and intended meaning, the authors manually verified the collected datasets. This verification process was crucial to confirm that the Arabic-English pairs were properly aligned and accurately conveyed the content of each other. 

 \subsection{Comparative Analysis of ATHAR and Other Arabic Datasets}
In this subsection, we analyze our dataset in comparison to existing classical and modern Arabic datasets, focusing on several linguistic measures: lexical diversity, stopword ratio, and the distribution of short versus long sentences, in addition to unique words count and dataset sizes.

We quantify lexical variety with the \emph{Measure of Textual Lexical Diversity}
\,(MTLD; \citealt{mccarthy2005assessment}).  The algorithm scans the text and
starts a new segment whenever the running type--token ratio (TTR) drops below a
fixed threshold; the MTLD score is the mean length of these segments.
Following \citeauthor{mccarthy2005assessment}, we set the threshold to
$\text{TTR}\le 0.75$, the lowest value that (i) aligns well with human
judgements of lexical variety, and (ii) remains stable for passages ranging from
1\,000 to 20\,000 tokens.

The stopword ratio was calculated by determining the occurrence of stopwords relative to the total word count in the datasets.
Short sentences were defined as any sentence containing 10 or fewer words, while long sentences are those with 30 or more words.

Before conducting the analysis, all datasets were standardized by removing redundant diacritics and letters, We chose to strip all diacritics to standardize the text format, since some source datasets were partially or fully diacritized while others were not. Furthermore, diacritics significantly expand the token space (e.g., distinguishing “\<كُتبٌ>”  from “\<كتب>”), complicating subword tokenization and increasing out-of-vocabulary rates. By using undiacritized text, we reduced preprocessing complexity and ensured consistent treatment across all corpora. As detailed in Table~\ref{tab:data_comparison}, the ATHAR dataset boasts one of the highest MTLD scores, suggesting that the text can sustain a high level of lexical diversity over a large number of words. This implies that the vocabulary is varied and the text does not quickly repeat words. Furthermore, our dataset maintains a balanced representation of both short and long sentences, providing a stark contrast to the variable sentence lengths found in other datasets.

\begin{table*}[ht]
\centering
\resizebox{\textwidth}{!}{%
\begin{tabular}{lcccccccc}
    \toprule
    \textbf{Dataset Attributes} & \textbf{ATHAR} & \textbf{Tanzil} & \textbf{Arabic PCD} & \textbf{KSUCCA} & \textbf{OPUS-100-ar-en} & \textbf{iwslt2017-ar-en} & \textbf{multiun-ar-en} \\
    \midrule
    Dataset size & 66K & 187K & 1.8M & 1.9M & 1M & 241K & 9.67M \\
    
    Unique words count & 138944 & 48104 & 720167 & 908771 & 370601 & 185390 & 841732 \\
    
    Lexical diversity (MTLD) & 55.63 & 101.31 & 11.86 & 40.87 & 17.46 & 34.12 & 70.10 \\ 
    
    Ratio of stopwords (\%) & 26.04 & 30.35 & 24.62 & 24.71 & 27.59 & 29.67 & 21.31 \\ 
    
    Average length of sentences & 20.78 & 34.35 & 9.26 & 25.33 & 8.39 & 13.86 & 22.89 \\ 
    
    Proportion of very short sentences (\%) & 24.06 & 11.18 & 76.57 & 41.28 & 79.81 & 45.98 & 23.07 \\ 
    
    Proportion of very long sentences (\%) & 23.11 & 47.53 & 0.00 & 24.44 & 4.71 & 7.04 & 26.61 \\ 
    
    \bottomrule
\end{tabular}%
}
\caption{Overview of Linguistic Characteristics in Arabic Language Datasets: Size, Diversity, and Sentence Metrics}
\label{tab:data_comparison}
\end{table*}

\section{Evaluating State-of-the-Art LLMs on the ATHAR Dataset}\label{sec:experiments}
In this section, we aim to evaluate the performance of state-of-the-art language models on classical Arabic translations using the ATHAR dataset. We selected four leading models for this analysis: GPT-4o, Llama-3 70B, Llama-3 8B, and Llama-2 7B.

Initially, we assessed the zero-shot capabilities of these models. Subsequently, we evaluated the Llama-3 8B and Llama-2 7B models under few-shot conditions. Finally, we focused on fine-tuning the Llama-3 8B model using two distinct methods: full fine-tuning, where all parameters of the model were adjusted, and LoRA~\cite{hu2021loralowrankadaptationlarge} parameter-efficient fine-tuning (PEFT), which only involved adjustments to a subset of newly added parameters. For LoRA, we adopted the default configuration provided in the Hugging Face PEFT documentation~\footnote{\url{https://huggingface.co/docs/peft/en/package_reference/lora}}: rank \(r = 8\), scaling factor \(\alpha = 8\), no dropout (\(0.0\)), no bias parameters trained (`bias = "none"`), and identity initialization (Kaiming-uniform for the A matrix and zeros for B). We utilized the HuggingFace Transformers~\footnote{\url{https://huggingface.co/docs/transformers/en/index}} library for full fine-tuning and inference of open-source models, and the OpenAI library~\footnote{\url{https://platform.openai.com/docs/libraries}} for GPT-4o. parameter-efficient Fine-tuning with LoRA was conducted using the HuggingFace PEFT library~\footnote{\url{https://huggingface.co/docs/peft/en/index}} implementation.

The objective of these comprehensive experiments is to maximize the potential of these models, understand performance variations under different settings, and explore how the ATHAR dataset can bridge existing performance gaps.

In the following subsections, we will detail the hyperparameters and metrics used in our experiments and analyze the results.
\subsection{Hyperparameters and Evaluation Metrics}
 \paragraph{Hyperparameters:} During inference, the generation decoding strategy involved setting the maximum number of new tokens to $2048$. Sampling strategies included Top-K and Top-P settings at $100$ and $0.95$, respectively, with a temperature parameter set at $0.3$. 

For fine-tuned models, specifically Llama-3 8B with full and LoRA tuning, training was implemented in an instruction input / response format. The input consisted of Arabic text, and the models were trained to generate the corresponding English translation as the response. The training dataset included $65k$ samples. The models were trained with precision $FP16$, with a learning rate of $5e-6$, adjusted using a linear scheduler over three epochs. The batch size was set at $16k$ tokens, which was achieved by accumulating gradients of four samples twice. An AdamW optimizer was utilized, with beta values of $0.90$ and $0.999$ for the first and second moment estimates, respectively.

The sentences were concatenated within the same source document, preserving the boundaries of the natural document. Each training example is a document fragment capped at 2,048 tokens ( 1300 Arabic + 700 English on average). The mean document length before splitting is 3 610 tokens ($\sigma = 2 140$), so ~40 \% of documents are split once, 8 \% twice, and the rest remain intact.

Regarding the prompt structures used in our experiments, Table~\ref{tab:models_prompts} details the specific prompt structures we utilized across zero-shot, few-shot, and fine-tuning settings.

\paragraph{Evaluation Metrics:} In assessing our models, we employed well-established metrics commonly used in translation evaluations: METEOR~\cite{banerjee-lavie-2005-meteor}, ROUGE-L~\cite{lin-2004-rouge}, and SacreBLEU~\cite{post-2018-call}. These metrics are all scored on a scale where higher values indicate better performance, though each has a different range. METEOR focuses on the alignment between the translation output and reference translations, considering synonymy and stemming. ROUGE-L measures the longest common subsequence, which is useful for evaluating the fluency of the text. SacreBLEU provides a consistent and comparable score across studies by standardizing the BLEU score calculation. Together, these metrics provide a comprehensive view of translation quality, covering aspects from accuracy to fluency. We utilized the HuggingFace Evaluate library~\footnote{\url{https://huggingface.co/docs/evaluate/en/index}} implementation for these metrics

\subsection{Results and Discussion}
\paragraph{Results:}The evaluation results, presented in Table~\ref{tab:performance_metrics}, highlight significant variances in the performance of the model in different settings. The GPT-4o model excelled in a zero-shot (ZS) setting, outperforming all other models with scores of $0.357$ in METEOR, $0.441$ in ROUGE-L, and $14.7$ in SacreBLEU. In contrast, the Llama-3 70B Instruct model, also evaluated in a zero-shot setting, registered slightly lower scores of $0.342$ in METEOR, $0.413$ in ROUGE-L and $13.0$ in SacreBLEU. This disparity might reflect differences in training regimes or underlying model architectures.

In the same zero-shot context, both the Llama-3 8B Instruct and Llama-2 7B models showed considerably lower performance in all metrics. These findings suggest inherent limitations in the zero-shot capabilities of these models for translation tasks.

Remarkable gains were observed with the \texttt{Llama-3-8B} model in the few-shot (FS) setting: using only three demonstrations, scores increased substantially to $0.174$ on METEOR, $0.167$ on ROUGE-L, and $0.971$ on SacreBLEU. These improvements highlight the strong in-context learning capabilities of the model. In contrast, \texttt{Llama-2-7B} exhibited only marginal improvements under few-shot evaluation. To test whether \texttt{Llama-2-7B}'s disparity was due to the number of examples, we performed a sweep over $k \in \{1,2,3,5\}$. As shown in Appendix~\ref{appendix_llama_fs_sweep} and Table~\ref{tab:llama2_fewshot}, performance in METEOR and ROUGE-L consistently remained below zero-shot levels, indicating that the limitation arises from model-specific sensitivity rather than the number of shots.

The Llama-3 8B model demonstrated further improvements after full fine-tuning, achieving a METEOR score of $0.275$, a ROUGE-L score of $0.336$ and a SacreBLEU score of $6.1$. Furthermore, the LoRA tuning method, which involves less extensive modifications, also yielded better results, with scores achieving $0.279$ on METEOR, $0.339$ on ROUGE-L and $8.8$ on SacreBLEU.

\paragraph{Discussion:} The results presented in Table~\ref{tab:performance_metrics} underscore the challenges faced by state-of-the-art LLMs when tasked with translating Classical Arabic to English. By providing state-of-the-art models with targeted training opportunities, the ATHAR dataset not only boosts model performance but also contributes significantly to the broader NLP community's understanding of and engagement with Classical Arabic. This dataset, therefore, holds substantial value, as it aids in developing more nuanced and capable translation systems.

\begin{table}[ht]
\centering
\resizebox{\columnwidth}{!}{%
\begin{tabular}{lccc}
    \toprule
    \textbf{Model} & \textbf{METEOR} $\uparrow$ & \textbf{ROUGE-L} $\uparrow$ & \textbf{SacreBLEU} $\uparrow$ \\
    \midrule
    GPT-4o + ZS (7th July 2024) & 0.357 & 0.441 & 14.7 \\
    Llama-3 70B Instruct + ZS & 0.342 & 0.413 & 13.0 \\
    Llama-3 8B Instruct + ZS & 0.115 & 0.068 & 0.3 \\
    Llama-2 7B + ZS & 0.116 & 0.099 & 0.3 \\ \midrule
    
    Llama-3 8B Instruct + FS3 & 0.174 & 0.167 & 1.0 \\
    Llama-2 7B + FS3 & 0.089  & 0.093 & 0.4 \\ \midrule

    Llama-3 8B + Full-Tuning & 0.275 & 0.336 & 6.1 \\
    Llama-3 8B + LoRA  & 0.279 & 0.339 & 8.8 \\
    
    \bottomrule
\end{tabular}%
}
\caption{Performance of State-of-the-Art LLMs on the Classical Arabic to English Translation Task. The table displays METEOR, ROUGE-L, and SacreBLEU scores for various models under different settings: zero-shot (ZS), few-shot with three samples (FS3), and fine-tuning (Full-Tuning \& LoRA) on a 1k test set.}
\label{tab:performance_metrics}
\end{table}

\begin{table*}[h!]
    \centering
    \small 
    \begin{tabular}{p{4.2cm} m{11cm}}
        \toprule
        \textbf{Model} & \textbf{Prompt} \\
        \midrule
        \makecell[l]{GPT-4o + ZS \\ 
        Llama-3 70B Instruct + ZS \\ 
        Llama-3 8B Instruct + ZS \\ 
        Llama-2 7B + ZS} 
        & Translate the following text from Classical Arabic to English\textbackslash nPlease return only the translated text without any introductions or additions:
        
        \{Arabic text\} \\ 
        \midrule
        \makecell[l]{Llama-3 8B Instruct + FS3 \\ 
        Llama-2 7B + FS3} 
        &
                        Translate the following Classical Arabic text into English. Follow the provided examples for consistency and accuracy.

                        Examples:
                        
                        Arabic: \begin{arabtext}باجر قَالَ ابْن دُرَيْد وَهُوَ صَنَمٌ كَانَ لِلأَزْدِ فِي الْجَاهِلِيَّةِ وَمَنْ جَاوَرَهُمْ من طَيء وقضاعة. كَانُوا يَعْبُدُونَهُ. بِفَتْحِ الْجِيمِ, وَرُبَّمَا قَالُوا بَاجِرُ بِكَسْرِ الْجِيمِ\end{arabtext}
                        English: According to Ibn Durayd, Bajar was an idol worshipped by the Azd tribe and neighboring tribes such as Tayyi’ and Quda’ah during the Jahiliyyah period. It is also pronounced Bajir.

                        Arabic: \begin{arabtext}(ذبل فِي آخر النُّسْخَة الَّتِي اعتمدتها فِي الطَّبْع)  الْيَعْبُوبُ صَنَمٌ لِجَدِيلَةِ طَيْءٍ. وَكَانَ لَهُمْ صَنَمٌ أَخَذَتْهُ مِنْهُمْ بَنُو أَسَدٍ فَتَبَدَّلُوا الْيَعْبُوبَ بَعْدَهُ. قَالَ عبيد فَتَبَدَّلُوا الْيَعْبُوبَ بَعْدَ إِلهِهِمْ ... صَنَمًا فَقَرُّوا يَا جديل واعذبوا (أَي لَا تَأْكُلُوا على ذَلِك وَلَا تشْربُوا)\end{arabtext}
                        English: Al-Ya'bub was the idol of the Jadilah tribe of Tayyi’. They initially had a different idol, but after it was taken by Banu Asad, they adopted Al-Ya'bub in its place. ‘Abid said they replaced their former god with Al-Ya'bub, and as a result, O Jadilah, abstain from food and drink.

                        Arabic: \begin{arabtext}فَلم يزل الْفلس يعبد حَتَّى ظهرت دَعْوَة النَّبِي عَلَيْهِ السَّلَام فَبعث إِلَيْهِ على ابْن أَبِي طَالِبٍ فَهَدَمَهُ وَأَخَذَ سَيْفَيْنِ كَانَ الْحَارِثُ بن أبي شمرٍ الغساني, ملك غَسَّان قَلَّدَهُ إِيَّاهُمَا, يُقَالُ لَهُمَا مِخْذَمٌ وَرَسُوبٌ(وَهُمَا السَّيْفَانِ اللَّذَانِ ذَكَرَهُمَا عَلْقَمَةُ بْنُ عَبْدَةَ فِي شِعْرِهِ). فَقَدِمَ بِهِمَا عَلِيُّ بْنُ أَبِي طَالِبٍ عَلَى النَّبِيِّ صَلَّى اللَّهُ عَلَيْهِ وَسَلَّمَ فَتَقَلَّدَ أَحَدَهُمَا ثُمَّ دَفَعَهُ إِلَى عَلِيِّ بْنِ أَبِي طَالِبٍ, فَهُوَ سَيْفُهُ الَّذِي كَانَ يَتَقَلَّدُهُ\end{arabtext}
                        English: Al-Fals continued to be worshipped until the advent of the Prophet. When ‘Ali ibn Abi Talib was sent to destroy it, he did so and took two swords, Mikhdham and Rasub, which had been given to Al-Fals by Al-Harith ibn Abi Shamir, the king of Ghassan. ‘Ali presented them to the Prophet, who kept one and returned the other to ‘Ali, who continued to wear it.

                        Translate the following:
                        
                        Arabic: \{Arabic text\}
                        
                        English: \\ 
        \midrule
        \makecell[l]{Llama-3 8B + Full-Tuning \\ 
        Llama-3 8B + LoRA} 
        & Translate the following input text from Classical Arabic to English, please return only the translated text without any introductions or additions.
                         
                        \#\#\# Input:
                        \{Arabic text\}
            
                        \#\#\# Response: \\ 
        \bottomrule
    \end{tabular}
    \caption{Prompt Structures Used and Their Corresponding Models in Zero-Shot (ZS), Few-Shot with Three Samples (FS3), and Full-Tuning Evaluation Experiments.}
    \label{tab:models_prompts}
\end{table*}

\section{Conclusion}\label{sec:conclusion}
To conclude, we introduce the ATHAR dataset, which enhances the existing corpus of Classical Arabic datasets by incorporating a broader range of topics. Our evaluation of the current status of LLMs underscores the critical need for the ATHAR dataset within the fine-tuning and training pipelines. More broadly, this need highlights the need for more comprehensive Classical Arabic datasets to improve the quality of translation systems in this domain. Future work will aim to expand the ATHAR dataset to include an even wider array of texts and topics, thus further enhancing translation quality.

\bibliography{custom}

\appendix
\section{ATHAR Data Sources}\label{appendix_athar_data_sources}
We drew 66 000 sentence pairs from 18 classical works spanning the 8$^{\text{th}}$--14$^{\text{th}}$ centuries
. The four largest sources ($\le$ 6000 sentence pairs each) are the History of al-Tabari, The Muqaddimah, The Book of Revenue, and The Travels of Ibn Battuta; complete counts appear in Table~\ref{tab:athar-sources}

\begin{table*}[t]
\centering
\caption{Primary sources in the ATHAR corpus, with century and topical domain.}
\label{tab:athar-sources}
\resizebox{\textwidth}{!}{%
\begin{tabular}{@{}l l l r@{}}
\toprule
\textbf{Title} & \textbf{Century} & \textbf{Topic} & \textbf{\# sentences} \\
\midrule
\emph{\<تاريخ الطبري> (History of al-Tabari)}                         & 10$^{\text{th}}$ & Universal history               & 9,591 \\
\emph{\<تُحفة النُّظار> (The Travels of Ibn Battuta)}                 & 14$^{\text{th}}$ & Travelogue                      & 9,591 \\
\emph{\<مقدمة ابن خلدون> (The Muqaddimah of Ibn Khaldun)}             & 14$^{\text{th}}$ & Historiography \& sociology     & 7,756 \\
\emph{\<الاموال> (The Book of Revenue)}                                &  9$^{\text{th}}$ & Economics \& public finance     & 7,420 \\
\emph{\<العقد الفريد> (The Unique Necklace)}                           & 10$^{\text{th}}$ & \emph{Adab} anthology           & 5,295 \\
\emph{\<المناظر> (The Optics)}                                          & 11$^{\text{th}}$ & Optics \& scientific method     & 4,148 \\
\emph{\<النوادر السلطانية و المحاسن اليوسفية> (The Sultan’s Anecdotes and Yusuf’s Merits)} & 12$^{\text{th}}$ & Biography                       & 4,086 \\
\emph{\<التصريف لمن عجز عن التأليف> (The Method of Healing)}           & 10$^{\text{th}}$ & Medical encyclopedia            & 3,164 \\
\emph{\<نشوار المحاضرة و أخبار المذاكرة> (Anecdotes of the Session and Stories of Recollection)} & 10$^{\text{th}}$ & Social \& cultural history      & 3,164 \\
\emph{\<القانون في الطب> (The Canon of Medicine)}                      & 11$^{\text{th}}$ & Medicine encyclopedia           & 2,507 \\
\emph{\<الاعتبار> (The Book of Reflection)}                            & 12$^{\text{th}}$ & Autobiographical narrative      & 2,286 \\
\emph{\<الرسالة> (The Epistle)}                                        &  9$^{\text{th}}$ & Islamic jurisprudence           & 2,001 \\
\emph{\<البخلاء> (The Book of Misers)}                                 &  9$^{\text{th}}$ & Satirical anecdotes (misers)    & 1,622 \\
\emph{\<نَهْجُ البَلاغَةِ> (The Path of Eloquence)}                      & 10$^{\text{th}}$ & Religious sermons               & 1,559 \\
\emph{\<فتوح الشام> (Fattouh al-Sham)}                                 &  9$^{\text{th}}$ & Military history                & 620 \\
\emph{\<الأخلاق والسير> (Ethics and Conduct)}                          & 11$^{\text{th}}$ & Ethics \& philosophy            & 603 \\
\emph{\<حي بن يقظان> (Hayy ibn Yaqdhan)}                                & 12$^{\text{th}}$ & Philosophical novel             & 435 \\
\emph{\<الاصنام> (The Book of Idols)}                                  &  9$^{\text{th}}$ & Pre-Islamic religion            & 195 \\
\midrule
\textbf{Total}                                                        & 18 works         &                              & 66,043 \\
\bottomrule
\end{tabular}
}
\end{table*}

\section{Data Samples}\label{appendix_data_sample}
Table~\ref{tab:samples_arabic} provides examples of classical Arabic text samples along with their English translations. Each row presents a segment of Arabic text followed by its corresponding English translation.
\begin{table*}[t]
  \centering
  \small
  \setlength{\extrarowheight}{2pt} 
  \begin{adjustbox}{max width=\textwidth}
    \begin{tabularx}{\textwidth}{@{} L L @{}}
      \toprule
      \textbf{Arabic} & \textbf{English} \\
      \midrule

      \begin{arabtext}ولم سموا البخل اصلاحا والشحّ اقتصادا، ولم حاموا على المنع، ونسبوه إلى الحزم؛ ولم نصبوا للمواساة، وقرنوها بالتضييع؟ ولم جعلوا الجود سرفا، والأثرة جهلا ؟ ولم زهدوا في الحمد، وقلّ احتفالهم بالذم\end{arabtext}
      &
      \begin{minipage}[t]{\hsize}\raggedright
        Why do they call avarice ‘ improvement’ and meanness ‘economy’? Why do they embrace cupidity and equate it with resolve while condemning generosity by likening it to waste? Why do they portray benevolence as extravagance and depict unselfishness as folly? Why are they so indifferent to the praise or blame of others
      \end{minipage} \\

      \addlinespace

      \begin{arabtext}وَكَانَ لِمُزَيْنَةَ صَنَمٌ يُقَالُ لَهُ نُهْمٌ. وَبِهِ كَانَتْ تُسَمَّى عَبْدُ نهمٍ. وَكَانَ سَادِنُ نهمٍ يُسمى خزاعى بْنَ عَبْدِ نهمٍ مِنْ مُزَيْنَةَ ثُمَّ مِنْ بَنِي عداءٍ\end{arabtext}
      &
      \begin{minipage}[t]{\hsize}\raggedright
        The Muzaynah had an idol called Nuhm. They used to name their children ‘Abd-Nuhm, after it. The cus- todian of Nuhm was called Khuza’i ibn-’Abd-Nuhm of the Muzaynah, and more specifically of the banu-’Ida
      \end{minipage} \\

      \addlinespace

      \begin{arabtext}وبلغنا أَنَّ رَسُولَ اللَّهِ عَلَيْهِ السَّلامُ قَالَ لَا تَذْهَبُ الدُّنْيَا حَتَّى تَصْطَكَّ أَلْيَاتُ نِسَاءِ دوسٍ عَلَى ذِي الْخَلَصَةِ يَعْبُدُونَهُ كَمَا كَانُوا يَعْبُدُونَهُ\end{arabtext}
      &
      \begin{minipage}[t]{\hsize}\raggedright
        We have been told that the Apostle of God once said, This world shall not pass away until the buttocks of the women of Daws wiggle again around dhu-al-Khalasah and they worship it as they were wont to do before Islam
      \end{minipage} \\

      \addlinespace

      \begin{arabtext}مثل استفراغ الْمَادَّة الفاعلة لوجع القولنج المحتبسة فِي لِيف الأمعاء وَإِمَّا سريع التَّأْثِير لكنه عَظِيم الغائلة مثل تخدير الْعُضْو الوجع فِي القولنج بالأدوية الَّتِي من شَأْنهَا أَن تفعل ذَلِك\end{arabtext}
      &
      \begin{minipage}[t]{\hsize}\raggedright
        Thus colic may be cured by purging the small intestine of the material giving rise to it, but this requires time. On the other hand one may give relief speedily, but only at the risk of worse harm in the end. Thus, it is possible to apply remedies which will in a case of colic at once make the painful part insensible
      \end{minipage} \\

      \addlinespace

      \begin{arabtext}وإن قوي الضوء الذي في الموضع، ثم لمح البصر ذلك المبصر من البعد البعيد الذي لمحه منه أولاً ولم يدرك حركته، فإنه قد يمكن أن يدرك حركته إذا لمحه والضوء الذي فيه قوي\end{arabtext}
      &
      \begin{minipage}[t]{\hsize}\raggedright
        If the light in that place becomes stronger and the eye glances at the object from that distance at which its motion was not perceived at first, sight will be able to perceive the strongly illuminated object
      \end{minipage} \\

      \addlinespace

      \begin{arabtext}فتفرق القوم عليهن وحدقوا بهن من كل جانب وراموا الوصول إليهن فلم يجدوا إلى ذلك سبيلا ولم تزل النساء لا يدنوا إليهن أحد من الروم إلا ضربن قوائم فرسه فإذا تنكس عن جواده بادرت النساء بالأعمدة فيقتلنه ويأخذن سلاحه\end{arabtext}
      &
      \begin{minipage}[t]{\hsize}\raggedright
        The Romans encircled them, but as soon as anyone came near, the women would break his horse’s legs with the pegs and when he thus fell down, would smash up his face
      \end{minipage} \\

      \bottomrule
    \end{tabularx}
  \end{adjustbox}
  \caption{Samples of classical Arabic texts and their English translations from classical sources.}
  \label{tab:samples_arabic}
\end{table*}


\section{Preprocessing: Flipped Cells in Data Collection}
\label{sec:flipped_cell_appendix}
During the scraping process, we encountered difficulties in extracting the English and Arabic texts from the containers (cells) because the Arabic texts were sometimes labeled as ``flex-right'' and English texts as ``flex-left'' in many instances, with the positions reversed in other cases. To address this, we counted the number of Arabic and English characters in each label and assigned the language based on the predominance of characters from either alphabet. Examples of such inconsistencies are provided below, where the labels for ``flex-right'' and ``flex-left'' are swapped, complicating the identification process:

\vspace{5pt}

\small 

\noindent \textbf{Example of Arabic Text on The Left and English Text on The Right:}
\vspace{4pt}

\noindent
\texttt{<div class="flex">}\\
\texttt{  <div class="flex-right">}\\
\texttt{    <span>"Farewell my brother, whom it was my duty to help. The blessings and the mercy of God upon you"</span>}\\
\texttt{  </div>}\\
\texttt{  <div class="flex-left">}\\
\begin{arabtext}
    والسلام عليك أيها الأخ المفترض إسعافه ورحمة الله وبركاته\\
\end{arabtext}
\noindent \texttt{</div>}\\
\texttt{</div>}\\

\vspace{5pt}

\noindent \textbf{Example of Arabic Text on The Right and English Text on The Left:}
\vspace{5pt}

\noindent
\texttt{<div class="flex">}\\
\texttt{  <div class="flex-right">}\\
\begin{arabtext}
    والمعافى أكثر من المبتلى؛ وليست الفوائد أقل من الحوائج\\
\end{arabtext}
\noindent \texttt{</div>}\\
\texttt{  <div class="flex-left">}\\
\texttt{    <span>"There are more healthy men than sick, after all, and good fortune is no less common than bad"</span>}\\
\texttt{  </div>}\\
\texttt{</div>}\\
\normalsize 

\section{Few-shot Sweep for \texttt{Llama-2-7B}}\label{appendix_llama_fs_sweep}

Table~\ref{tab:llama2_fewshot} reports the performance of \texttt{Llama-2-7B} across a range of few-shot settings. The goal of this sweep is to investigate whether the lack of improvement compared to zero-shot evaluations is attributable to model-specific limitations or to sensitivity with respect to the number of shots. The results indicate that performance does not consistently improve with additional demonstrations, suggesting that the observed sensitivity is not primarily due to the number of shots.
\begin{table}[ht]
\centering
\resizebox{\columnwidth}{!}{%
\begin{tabular}{lccc}
\toprule
Model & \multicolumn{3}{c}{\texttt{Llama-2-7B}} \\
\cmidrule(lr){2-4}
Few-shot ($k$) & METEOR $\uparrow$ & ROUGE-L $\uparrow$ & SacreBLEU $\uparrow$ \\
\midrule
1 & 0.050  & 0.077 & 0.4 \\
2 & 0.064  & 0.061 & 0.6 \\
3 & 0.089  & 0.093 & 0.4 \\
5 & 0.065  & 0.065 & 0.4 \\
\bottomrule
\end{tabular}%
}
\caption{Few-shot results for \texttt{meta-llama/Llama-2-7b-hf} with $k \in \{1,2,3,5\}$. Performance is measured using METEOR, ROUGE-L, and SacreBLEU.}
\label{tab:llama2_fewshot}
\end{table}

\end{document}